\documentclass[10pt,journal,compsoc,two column]{IEEEtran}%
\usepackage{amsfonts}
\usepackage{amsmath}
\usepackage{graphicx}
\usepackage{subfigure}
\usepackage{multirow}
\usepackage{amssymb}
\usepackage{epstopdf}
\usepackage{placeins}
\usepackage{booktabs}
\usepackage{algorithmic}
\usepackage{graphicx}
\usepackage{placeins}
\usepackage{epstopdf}
\usepackage{lineno,hyperref}
\usepackage[nocompress]{cite}
\usepackage{cite}
\usepackage{algorithm}
\usepackage{algorithmic}%
\providecommand{\U}[1]{\protect\rule{.1in}{.1in}}
\ifCLASSOPTIONcompsoc
\else
\fi
\graphicspath{{figures/}}

\makeatother
\begin{document}

\title{Learning to embed semantic similarity for joint image-text retrieval}
\author{~Noam Malali and Yosi~Keller~
\IEEEcompsocitemizethanks{\IEEEcompsocthanksitem  N. Malali \& Y. Keller are with the Faculty of Engineering, Bar-Ilan University,
E-mail:yosi.keller@gmail.com}\thanks{Manuscript received April 19, 2005;
revised August 26, 2015.}}
\maketitle

\begin{abstract}
We present a deep learning approach for learning the joint semantic embeddings
of images and captions in a Euclidean space, such that the semantic similarity
is approximated by the $L_{2}$ distances in the embedding space. For that, we
introduce a metric learning scheme that utilizes multitask learning to learn
the embedding of \textbf{identical} semantic concepts using a center loss. By
introducing a differentiable quantization scheme into the end-to-end trainable
network, we derive a semantic embedding of semantically \textbf{similar}
concepts in Euclidean space. We also propose a novel metric learning
formulation using an adaptive margin hinge loss, that is refined during the
training phase. The proposed scheme was applied to the MS-COCO, Flicke30K and
Flickr8K datasets, and was shown to compare favorably with contemporary
state-of-the-art approaches.

\end{abstract}







\markboth{IEEE TRANSACTIONS ON IMAGE PROCESSING,~VOL.~, NO.~,
~}{Shell \MakeLowercase{\textit{et al.}}: Bare Demo of IEEEtran.cls
for IEEE Journals}


\IEEEpeerreviewmaketitle


\section{Introduction}

\label{sec:Introduction}

\begin{figure}[tbh]
\centering\includegraphics[width=1.0\linewidth]{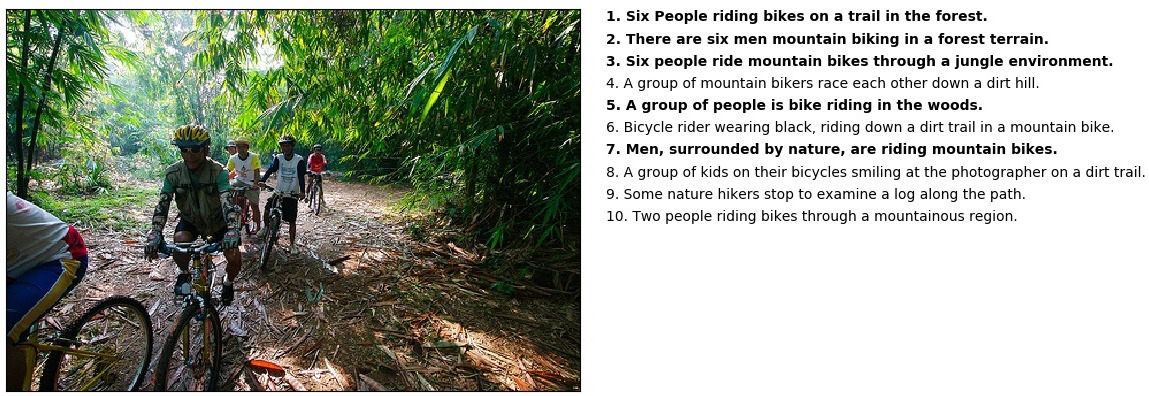}\caption{Image
caption retrieval (image to text), using the Flickr30K dataset
\cite{flickr30k}. The embedding of an image is compared with the embedding of
all sentences in the dataset, and the closest ten captions are presented. The
sentences marked in bold are the ground truth matches having
\textit{identical} semantic content, in contrast to the other sentences having
\textit{similar} content.}%
\label{fig:i2t_example_intro}%
\end{figure}

The joint embedding of text and image data is a contemporary research theme
paving the way for novel applications such as Visual Question Answering
\cite{VQA}, auto-caption generation \cite{W17-3506}, and image-to-text
retrieval \cite{Kiros2014}\cite{feifeiKarpathy}. Figure
\ref{fig:i2t_example_intro} depicts image-to-text retrieval where a query
image is embedded in a joint image-text space, and the chosen caption is the
one closest to the image in terms of their $L_{2}$ distance, in the joint
embedding space. Similarly, given a caption, the joint embedding space allows
text-to-image retrieval, where the closest image is retrieved. The computation
of the joint embedding space is commonly formulated as a Metric Learning
problem using training sets of corresponding images and captions.

Let the image and caption, be denoted by $\mathbf{x}$ and $\mathbf{y}$,
respectively. We aim to compute their corresponding joint embedding
$\mathbf{\hat{x},\hat{y}}\in%
\mathbb{R}
^{d}$, using the mapping functions $\mathbf{H}_{i}$ and $\mathbf{H}_{t},$
respectively, such that
\begin{equation}
\mathbf{\hat{x}}_{n}=\mathbf{H_{i}}(\mathbf{x}_{n}),\widehat{\mathbf{y}}%
_{n}=\mathbf{H_{t}}(\mathbf{y}_{n})
\end{equation}

The mappings $\left\{  \mathbf{H}_{i},\mathbf{H}_{t}\right\}  $ are learned
using a supervised scheme, given a training set of $S=\left\{  \mathbf{x}%
_{i}\mathbf{,y}_{i}\right\}  _{1}^{N}$ samples, and corresponding binary
labels. In practice, $S$ only consists of semantically identical pairs, while
all other pairs are considered dissimilar. $\left\{  \mathbf{H}_{i}%
,\mathbf{H}_{t}\right\}  $ are learned to optimize a distance metric, such as
$L_{1}$ and $L_{2}$, over the training set..

Metric learning was applied in a plethora of problems and classical
formulations \cite{Weinberger:2005:DML:2976248.2976433,NIPS2004_2566}. In this
work, we relate to metric learning in the context of joint image-text
embeddings, where several schools of thought were studied. Some schemes
utilize only similar pairs to increase the correlation between their
embeddings, based on variants of Canonical Correlation Analysis (CCA)
\cite{cca} such as KCCA \cite{kcca}, and their extensions using Convolutional
Neural Networks (CNNs), such as the 2WayNet \cite{2WayNetwork}, DCCA
\cite{deepcca} and CorrNet \cite{corrnet}.

Other approaches, such as Visual Semantic Embedding (VSE++) \cite{vse++}, Dual
Attention Network (DAN) \cite{dan}, sm-LSTM \cite{sm-lstm} and Embedding
Network \cite{emeddingNetwork} utilize both similar and dissimilar pairs to
minimizes the distance of the similar pairs in the embedding space, while
maximizing the distance of the dissimilar ones. In the Contrastive loss a
hinge loss is applied to the distance metric of dissimilar samples%
\begin{equation}
\mathcal{L}_{p}=\sum_{n=1}^{N}D(\mathbf{\hat{x}}_{n},\mathbf{\hat{y}}_{n}%
^{+})+\sum_{n=1}^{N}\left[  m-D(\mathbf{\hat{x}}_{n},\mathbf{\hat{y}}_{n}%
^{-})\right]  _{+} \label{eq:loss_pairwise}%
\end{equation}
where $\left\{  \mathbf{\hat{x}}_{n},\mathbf{\hat{y}}_{n}^{+}\right\}  $ and
$\left\{  \mathbf{\hat{x}}_{n},\mathbf{\hat{y}}_{n}^{-}\right\}  $ are a pair
of embeddings of similar and dissimilar samples, respectively, and $\left[
x\right]  _{+}=\max\left(  x,0\right)  .$ $m$ is predefined, and it is common
to set $m=1$ for $L_{2}$ normalized embeddings. A Triplet Loss \cite{4408839}
optimizes the difference in distances between similar $\left\{  \mathbf{\hat
{x}}_{n},\mathbf{\hat{y}}_{n}^{+}\right\}  $ and dissimilar $\left\{
\mathbf{\hat{x}}_{n},\mathbf{\hat{y}}_{n}^{-}\right\}  $ pairs%
\begin{equation}
\mathcal{L}_{T}=\sum_{n=1}^{N}\left[  D(\mathbf{\hat{x}}_{n},\mathbf{\hat{y}%
}_{n}^{+})-D(\mathbf{\hat{x}}_{n},\mathbf{\hat{y}}_{n}^{-})+m\right]  _{+},
\label{equ:triplet loss}%
\end{equation}
where $\left\{  \mathbf{\hat{x}}_{n},\mathbf{\hat{y}}_{n}^{+},\mathbf{\hat{y}%
}_{n}^{-}\right\}  $ is a triplet of samples such that $\mathbf{\hat{x}}_{n}$
is an image embedding, while $\mathbf{\hat{y}}_{n}^{+}$ and $\mathbf{\hat{y}%
}_{n}^{-}$ are semantically similar and dissimilar captions, respectively.

Recent works \cite{Huang2018LearningSC,SCAN,FRCNN} applied bottom-up attention
schemes, where the image is semantically parsed, and the resulting ROIs are
matched to the captions. The ROIs are detected by Faster RCNN
\cite{NIPS2015_5638}. The number of dissimilar pairs far exceeds that of
similar ones, and as most dissimilar pairs are easy to detect, they prove
uninformative for training. Thus, informative dissimilar samples denoted as
\textit{hard negatives} have to be chosen explicitly.

In this work, we formulate the image-text embedding as the embeddings of
semantically \textit{identical} and \textit{similar} images and captions. The
sets of semantically \textit{identical} images and captions are given by the
training sets $S=\left\{  \mathbf{s}_{n}\right\}  _{1}^{N}=\left\{
\mathbf{x}_{n},\mathbf{y}_{n}^{1},...,\mathbf{y}_{n}^{K}\right\}  _{1}^{N}$
\cite{HodoshFlickr8k,flickr30k,mscoco}. Each subset $\mathbf{s}_{n}$ consists
of an image $\mathbf{x}_{n}$ and $K$ corresponding (semantically identical)
captions $\mathbf{y}_{n}^{1},...,\mathbf{y}_{n}^{K}$. We propose to quantize
the semantic embeddings using an end-to-end deep learning scheme, to derive an
embedding that relates \textit{similar} images and captions, that are not
given by the training sets. The joint embedding is computed by applying a
network consisting of two embedding branches. The first computes the image
representation using convolution layers, while the second computes the text
embedding using word embedding and GRU layers, as depicted in Fig.
\ref{fig:CNN outline}. As identical samples are a subset of the similar ones,
we propose a novel approach to separately induce the similarity of identical
and similar images and captions, such that the similarity of identical samples
is increased compared to that of the similar ones.

Thus, the proposed CNN is trained using multitask learning. We apply a the
proposed Semantic Center Loss to compute the semantic embedding and induce the
similarity of \textit{similar} samples, and a novel formulation of the Triplet
Loss, using an adaptive-margin, to induce the similarity of \textit{identical
}samples. Cross-entropy losses are applied to each of the image and text
embedding branches separately, to improve their convergence. The proposed
semantic embedding and Triplet Loss formulation can be used with \textit{any}
CNN backbone implementing the image and text embeddings.

In particular, we propose the following contributions:

\textbf{First}, we present a CNN-based joint image-text embedding scheme to
learn a semantic representation using training sets of images and captions
encoding \textit{identical} semantic notions.

\textbf{Second}, by quantizing the resulting semantic embeddings using an
end-to-end deep learning approach, we agglomerate the semantic representation
to derive an embedding that encodes semantic \textit{similarity} in terms of
$L_{2}$ distances in the embedding space.

\textbf{Third}, we propose a novel Triplet Loss formulation that utilizes a
hinge loss with adaptive margins that is shown to improve the embedding accuracy.

\textbf{Last}, the proposed scheme is shown to compare favorably with
contemporary state-of-the-art approaches when applied to the retrieval of
images given their corresponding captions, and vice versa.

\section{Background}

\label{sec:Background}

Metric learning is a fundamental topic in machine learning applied in a
plethora of problems and classical formulations
\cite{Weinberger:2005:DML:2976248.2976433,NIPS2004_2566}, and was used in
joint image-text embeddings.

Klein et al. \cite{cca} applied Canonical Correlation Analysis (CCA), to
compute a linear projection of two data views into a joint space that
maximizes the correlation between their embeddings. CCA utilizes only positive
pairs of samples and has been applied by Li et al. \cite{cca_face} to face
recognition, and by Klein et al. \cite{cca} for relating text to images. A
Deep canonical correlation analysis (DCCA) was proposed by Andrew et al.
\cite{deepcca}, where CCA was used as a loss function.

Other deep learning schemes such as Correlation Neural Networks (CorrNet) by
Chandar et al. \cite{corrnet} and DCCAE by Wang et al. \cite{DCCAE} utilized
an autoencoder architecture with two input and output views. Contemporary Deep
CCA methods optimize the CCA loss on top of a CNN. Thus, Eisenschtat and Wolf
\cite{2WayNetwork} introduced the 2Way Network architecture consisting of a
bidirectional CNN\ that was optimized using an $L_{2}$ loss, to match the two
data sources. Another widely used approach is the ranking hinge loss, which
utilizes positive/similar and negative/dissimilar data pairs, to learn a
representation in which the positive pairs are closer than negative ones. A
pairwise hinge ranking loss was applied by Chechik et al. \cite{Chechik:2010}
for learning image similarity and similarity-preserving hash functions.
Similarly, Norouzi and Fleet \cite{Norouzi} studied the embedding of high
dimensional data as binary codes.

Pairwise losses were applied to image-text metric learning by Weston et al.
\cite{Weston} for the annotation of large scale image datasets. A ranking
hinge loss for deep visual semantic embedding was applied by Frome et al.
\cite{DeViSE} to improve the identification, and recognition of visual
objects. Kiros et al. \cite{Kiros2014} proposed to apply the triplet ranking
loss to visual semantic embedding and image captioning. The same loss was also
used by Karpathy et al. \cite{feifeiKarpathy} and Socher et al. \cite{socher}
to generate natural language captioning of images. The learning of similarity
embedding using CNN was suggested by Wang et al. \cite{emeddingNetwork} by
applying a maximum-margin triplet ranking loss. Hard Mining was shown to
improve CNN training in general, and metric learning in particular by Faghri
et al. \cite{vse++}. Image-text similarity was formulated as a binary
classification problem by Jabri et al. \cite{JabriJM16} for Visual Question
Answering (VQA), to predict the matching of images and questions. The same
loss was used in a zero-shot formulation by Ba et al. \cite{Ba} to match
images and captions using a CNN consisting of two CNN subnetworks. The
accuracy of metric learning-bases was improved by Wu et al.
\cite{Sampling-Matters} by introducing an adaptive margin into the Hinge loss
within the Contrastive loss, as in Eq. \ref{eq:loss_pairwise}. The adaptive
margin is optimized during the training phase. Similarly, Chen
\cite{Beyond-Triplet} proposed an adaptive margin based on the average
distance between the positive and negative training samples, to improve the
proposed quadruplet metric learning scheme. A class-adaptive margin loss was
proposed by Li et al. \cite{Li_2020_CVPR}, where semantic similarity between
each pair of classes was analyzed to separate samples from similar classes in
the embedding domain. The proposed margin was applied to improve the
generalization of meta-learning in few-shot learning problems.

Attention-bases approaches \cite{Huang2018LearningSC,FRCNN,SCAN} applied
Bottom-up visual parsing to match images and captions, where the set of
semantic concepts is represented by a vocabulary of words derived by analyzing
the captions in the training set. The training images were visually parsed by
extracting multiple ROIs from each image using the Region Proposal Network
(RPN) of the Faster R-CNN \cite{NIPS2015_5638}, and associating semantic
concepts as attributes per ROI. Huang et al. \cite{Huang2018LearningSC}
utilize a multi-label CNN to predict the semantic concepts per ROI, that are
merged, and fused with a global image context, resulting in a joint embedding.
Lee et al. \cite{SCAN} applied stacked cross attention to match the ROIs
estimated by the bottom-up image parsing, to words, to derive a one-to-one
matching scheme without computing a joint embedding in an $L_{2}$ space. Li et
al. \cite{FRCNN} extended these schemes by applying Graph Convolutional
Networks (GCN) to the embeddings of the extracted ROIs, to derive features
with semantic relationships and the multiple ROI\ embeddings were fused using
an RNN. This approach achieved state-of-the-art accuracy. In addition to a
training set of semantically identical images and captions (MS-COCO,
Fliker30K), such schemes require a large-scale annotated set of instances
(`building,' `giraffe,' etc.) and attributes (`furry,' `fast,' etc.) appearing
in the images, to train the Faster RCNN \cite{NIPS2015_5638}. The
\textit{Support-set} work \cite{supportset} by Patrick et al. is of particular
interest, as it studies the joint embedding of video sequences and text.
Patrick et al. propose a center loss with soft assignments, whose centers are
chosen from the embeddings of the frames of the video sequence. Thus, the
embedding of each caption is a linear combination of the embeddings of the
Support-set video frames.

\section{Joint Semantic Embedding of Images and Captions}

\label{chap:OurModel}

Let $S=\left\{  \mathbf{s}_{n}\right\}  _{1}^{N}=\left\{  \mathbf{x}%
_{n},\mathbf{y}_{n}^{1},...,\mathbf{y}_{n}^{K}\right\}  _{1}^{N}$ be a set of
subsets $\mathbf{s}_{n}$ each consisting of an image $\mathbf{x}_{n}$ and $K$
captions $\left\{  \mathbf{y}_{n}^{1},...,\mathbf{y}_{n}^{K}\right\}  $. We
aim to compute the joint embeddings of $S$
\begin{equation}
\widehat{S}=\left\{  \mathbf{\hat{x}}_{n},\widehat{\mathbf{y}}_{n}%
^{1},...,\widehat{\mathbf{y}}_{n}^{K}\right\}  _{1}^{N},\mathbf{\hat{x}}%
_{n},\widehat{\mathbf{y}}_{n}^{k}\in%
\mathbb{R}
^{d} \label{eq:subset}%
\end{equation}
such that the embeddings $\widehat{S}$ encode the semantic similarities of
$S,$ by minimizing the $L_{2}$ distances between semantically similar images
and captions, and maximizing the distances between semantically dissimilar ones.

Sets of semantically \textit{identical} images and captions $S$ are given in
the Flickr8K \cite{HodoshFlickr8k}, Flickr30K \cite{flickr30k}, and MS-COCO
\cite{mscoco} training sets. Each of these training datasets $S$ consists of
subsets $\mathbf{s}_{n}$ containing a single image and $K=5$ corresponding
captions that encode an \textit{identical }semantic concept.

\subsection{Network Backbone}

\begin{figure}[tbh]
\centering\includegraphics[width=1.0\linewidth]{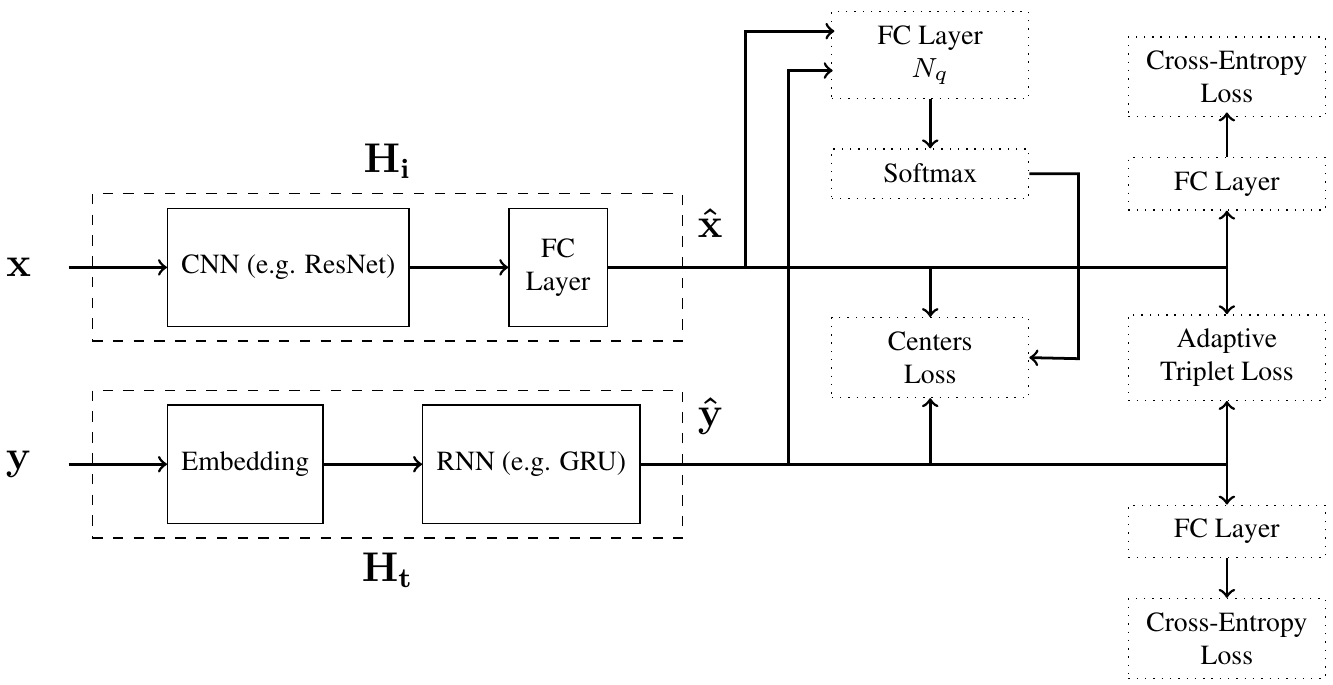}\caption{The
proposed joint embedding scheme. The outputs of the image and text branches
are $\mathbf{\hat{x}}$ and $\widehat{\mathbf{y}}$, respectively. We used
multiple different backbones. The backbone shown follows
\cite{2WayNetwork,vse++}. The subnetwork to the right of $\mathbf{\hat{x}}$
and $\widehat{\mathbf{y}}$ is common to \textit{all} our architectures, and is
the \textit{core} of our approach. $\mathbf{\hat{x}}$ and $\widehat
{\mathbf{y}}$ are connected to separate cross-entropy losses labeled by their
given set $\mathbf{s}_{n}$. A triplet hinge loss with adaptive margins, as in
Section \ref{section:adaptive_margin}, is used to merge the embeddings. The
\textit{unquantized} embedding is computed by directly connecting
$\mathbf{\hat{x}}$ and $\widehat{\mathbf{y}}$ to the Center Loss as in Section
\ref{section:constrained_loss}, to compute the semantic centers $\left\{
\mathbf{c}_{n}\right\}  _{1}^{N}$. For the \textit{quantized} embedding as in
Section \ref{section:quantized constrained_loss}, $\mathbf{\hat{x}}$ and
$\widehat{\mathbf{y}}$ are connected to the \textquotedblleft$N_{q\text{ }}%
$FC\textquotedblright\ and Softmax layers to compute $N_{q}\ll N$ embedding
centers $\left\{  \widehat{\mathbf{c}}_{n}\right\}  _{1}^{N_{q}}$.}%
\label{fig:CNN outline}%
\end{figure}

The outline of the proposed CNN\ is shown in Fig. \ref{fig:CNN outline}. The
proposed scheme consists of image and text embedding branches, $\mathbf{H}%
_{i}$ and $\mathbf{H}_{t}$, respectively, that compute the image and text
embeddings $\mathbf{\hat{x}}_{n}$ and $\widehat{\mathbf{y}}_{n}$. We used
\textit{multiple} different backbone architectures, to compare against recent
results as reported in Section \ref{sec:Experiments}. All different backbones
compute (differently) the image and text embeddings $\mathbf{\hat{x}}_{n}$ and
$\widehat{\mathbf{y}}_{n}$, as in Fig. \ref{fig:CNN outline}, while the rest
of the proposed network, that is the core of our contribution, is detailed in
Section \ref{section:semantic embeddings}.

Figure \ref{fig:CNN outline} shows a backbone following 2WayNet
\cite{2WayNetwork} and VSE++\cite{vse++}, where the input image $\mathbf{x}%
_{n}$ is embedded by VGG19 or ResNet152 CNNs. An FC layer adjusts the output
dimensionality of the image embeddings $\mathbf{\hat{x}}_{n}$ to that of the
text embeddings $\widehat{\mathbf{y}}_{n}$. The text embedding is computed by
an embedding layer followed by a GRU layer. We also used the Attention-based
backbone proposed by the (previous) SOTA VSRN approach \cite{FRCNN} by Li et
al. 

\subsection{Semantic Embedding Subnetwork}

\label{section:semantic embeddings}

Given the image and text embeddings $\mathbf{\hat{x}}_{n}$ and $\widehat
{\mathbf{y}}_{n}$, as in\ Fig. \ref{fig:CNN outline}, the Semantic embedding
subnetwork (to the right of $\mathbf{\hat{x}}_{n}$ and $\widehat{\mathbf{y}%
}_{n}$ in\ Fig. \ref{fig:CNN outline}) is shared by \textit{all} of our
proposed schemes and is the \textit{core} of our contribution. The semantic
embedding of identical subsets $\mathbf{s}_{n}$ is learned by applying the
Semantic Center Loss $\mathcal{L}_{C}$, detailed in Section
\ref{section:constrained_loss} to minimize the inner-set variance of the
embedding $\widehat{s}_{n}$ of the subsets $s_{n}$. Different training subsets
$\left\{  \mathbf{s}_{n}\right\}  $ might encode \textit{similar} (not
identical) semantic concepts. But, the semantic similarity is not given by the
training sets. To compute an embedding that encodes semantic similarity, we
propose to quantize the embeddings$\left\{  \widehat{s}_{n}^{i}\right\}  $
using $N_{q}\ll N$ semantic centers as discussed in Section
\ref{section:quantized constrained_loss}. This allows to agglomerate $N$
identical concepts to $N_{q}$ similar concepts.

We also apply a novel formulation of the Triplet Hinge Loss $\mathcal{L}_{T}$
detailed in Section \ref{section:adaptive_margin}, that utilizes adaptive
hinge loss margins. The margins are updated during the training and allow
better separation (in the $L_{2}$ sense) of the different semantic concepts.

Last, we apply two separate cross-entropy (CE) losses, $\mathcal{L}_{CE}^{I}$
and $\mathcal{L}_{CE}^{T}$, for the image and text embeddings, respectively.
These losses classify each embedding $\mathbf{\hat{x}}_{n}\in\mathbf{s}_{n}$
or $\widehat{\mathbf{y}}_{n}\in\mathbf{s}_{n}$ with respect to the index $n$
of its subset $\mathbf{s}_{n}$. As $S=\left\{  \mathbf{s}_{n}\right\}
_{1}^{N},$ there are $N$ such classes. The overall loss is the unweighted sum
of all losses above%
\begin{equation}
L=\mathcal{L}_{C}+\mathcal{L}_{T}+\mathcal{L}_{CE}^{I}+\mathcal{L}_{CE}^{T}.
\label{equ:loss}%
\end{equation}
We experimented in weighing the different loss, but the accuracy was not improved.

\subsection{Semantic Center Loss}

\label{section:constrained_loss}

Each subset $s_{n}=\left\{  \mathbf{x}_{n},\mathbf{y}_{n}^{1},...,\mathbf{y}%
_{n}^{K}\right\}  $ in the given training sets
\cite{HodoshFlickr8k,flickr30k,mscoco} consists of semantically
\textit{identical} entries. Let $\mathbf{c}_{n}\in%
\mathbb{R}
^{d}$ be the embedding (semantic center) corresponding to the subset $s_{n}$.
$\mathbf{c}_{n}$ and $s_{n}$ are apriori assigned and we have $C=\left\{
\mathbf{c}_{n}\right\}  _{1}^{N}$ semantic centers that are learned in the
training phase. The \textit{same} embedding center $\mathbf{c}_{n}$ is applied
to the embeddings of \textit{both} image and captions training entries
$\left\{  \mathbf{x}_{n},\mathbf{y}_{n}^{1},...,\mathbf{y}_{n}^{K}\right\}
\in s_{n}$, as both are given to be semantically identical.

We propose to minimize the Center Loss \cite{CenterLoss} with respect to both
images and captions
\begin{equation}
\mathcal{L}_{C}=\sum_{i}\left(  \Vert\mathbf{\hat{x}}_{i}-\mathbf{c}%
_{(\widehat{\mathbf{x}}_{i},\widehat{\mathbf{y}}_{i})}\Vert_{2}^{2}%
-\delta\right)  _{+}+\left(  \Vert\widehat{\mathbf{y}}_{i}-\mathbf{c}%
_{(\widehat{\mathbf{x}}_{i},\widehat{\mathbf{y}}_{i})}\Vert_{2}^{2}%
-\delta\right)  _{+} \label{equ:semantic loss}%
\end{equation}
where $\mathbf{c}_{(\widehat{\mathbf{x}}_{i},\widehat{\mathbf{y}}_{i})}$ is
the center corresponding to both $\mathbf{\hat{x}}_{i}$ and $\widehat
{\mathbf{y}}_{i}$, respectively. During training, the proposed CNN is given
triplets of input images and captions used for the metric learning detailed in
Section \ref{section:adaptive_margin}, and the Center Loss $\mathcal{L}_{C}$
is minimized for each image and caption training sample.

\subsection{Quantized Semantic Center Loss}

\label{section:quantized constrained_loss}

The Center Loss in Eq. \ref{equ:semantic loss} minimizes the inner variability
of each subset $s_{n}\in S$. But, the resulting embedding does not minimize
the $L_{2}$ distance between the embeddings of semantically \textit{similar
(}but\textit{\ }not identical) images and corresponding captions. Moreover,
with the use of large training sets, such as the MS-COCO \cite{mscoco}
dataset, the number of subsets $\left\vert S\right\vert =N$ and corresponding
centers $\mathbf{c}_{n}$ increases significantly, to $N>100K$, implying that
many of the semantic concepts encoded by the training sets $S$ are similar.
Thus, the semantic centers $\left\{  \mathbf{c}_{n}\right\}  _{1}^{N},$
computed in Section \ref{section:constrained_loss}, can be quantized to derive
an embedding that encodes \textit{semantic similarity}, such that the samples
in semantically similar sets $s_{n}$ will have a similar embedding in the
$L_{2}$ metric.

Hence, we propose to quantize the semantic centers $\left\{  \mathbf{c}%
_{n}\right\}  _{1}^{N}$ to $C_{q}=\left\{  \widehat{\mathbf{c}}_{n}\right\}
_{1}^{N_{q}}$, $N_{q}\ll N,$ by applying a \textit{soft} differentiable
quantization scheme, where a Softmax layer is utilized to assign soft labels
\cite{netvlad}. The soft assignments $\mathbf{w}_{\mathbf{\hat{x}}_{i}%
},\mathbf{w}_{\widehat{\mathbf{y}}_{i}}\in%
\mathbb{R}
^{N_{q}}$ are computed by applying the Softmax layer to $\mathbf{\hat{x}}_{i}$
(image) and $\widehat{\mathbf{y}}_{i}$ (text), respectively.

Given the soft assignments to the quantized semantic centers $C_{q}$, the
centers are learned using an extension of the Semantic Center Loss
\cite{DBLP:journals/corr/abs-1708-02551}.
\begin{multline}
\mathcal{L}_{C}=\sum_{i}\sum_{j=1}^{^{N_{q}}}\mathbf{w}_{\mathbf{\hat{x}}_{i}%
}\left(  \mathbf{c}_{j}\right)  \left(  \Vert\mathbf{\hat{x}}_{i}%
-\widehat{\mathbf{c}}_{j}\Vert_{2}^{2}-\delta\right)  _{+}%
\label{eq:loss_center_modified_1}\\
+\sum_{i}\sum_{j=1}^{^{N_{q}}}\mathbf{w}_{\widehat{\mathbf{y}}_{i}}\left(
\mathbf{c}_{j}\right)  \left(  \Vert\widehat{\mathbf{y}}_{i}-\widehat
{\mathbf{c}}_{j}\Vert_{2}^{2}-\delta\right)  _{+}\\
+\alpha\sum_{k_{1},k_{1}}^{N_{q}}\left(  2\delta-\left\Vert \widehat
{\mathbf{c}}_{k_{1}}-\widehat{\mathbf{c}}_{k_{2}}\right\Vert _{2}^{2}\right)
_{+},
\end{multline}
where $\mathbf{w}_{\mathbf{\hat{x}}_{i}}\left(  \mathbf{c}_{j}\right)  $ and
$\mathbf{w}_{\widehat{\mathbf{y}}_{i}}\left(  \mathbf{c}_{j}\right)  $ are the
soft assignment weights of the sample $\mathbf{\hat{x}}_{i}$ and
$\widehat{\mathbf{y}}_{i}$ to the center $\mathbf{c}_{j}$, and $\alpha
>0\,\ $is a weight term that was set to $\alpha=1$. Note that as
$\mathbf{w}_{\mathbf{\hat{x}}_{i}}$ and $\mathbf{w}_{\widehat{\mathbf{y}}_{i}%
}$ are the outputs of a Softmax layer, we have that
\begin{equation}
\sum_{j=1}^{^{N_{q}}}\mathbf{w}_{\mathbf{\hat{x}}_{i}}\left(  \mathbf{c}%
_{j}\right)  =1,\forall\mathbf{\hat{x}}_{i}%
\end{equation}
and%
\begin{equation}
\sum_{j=1}^{^{N_{q}}}\mathbf{w}_{\widehat{\mathbf{y}}_{i}}\left(
\mathbf{c}_{j}\right)  =1,\forall\widehat{\mathbf{y}}_{i}.
\end{equation}

Thus, Eq. \ref{eq:loss_center_modified_1} implies that both $\mathbf{\hat{x}%
}_{i}$ and $\widehat{\mathbf{y}}_{i}$ are reconstructed as a linear
combination of the quantized centers $C_{q}=\left\{  \widehat{\mathbf{c}}%
_{n}\right\}  _{1}^{N_{q}}$.

In this formulation, the Center Loss in Eq. \ref{equ:semantic loss} is
extended by adding a term that pushes the centers away from each other, to
improve the separability of the embeddings of centers that are semantically
dissimilar. As the training sets only provide semantically identical subsets
$s_{n}$, it is apriori unknown whether different training subsets $s_{n_{1}}%
$and $s_{n_{2}}$ $n_{1}\neq n_{2}$ are similar or dissimilar. Thus, the
additional term in Eq. \ref{eq:loss_center_modified_1} could not be used
without applying the differentiable semantic quantization that introduces the
notion of semantic dissimilarity. Equation \ref{eq:loss_center_modified_1}
does not distinguish between similar and identical samples, as similar samples
are a superset of the identical ones. In order to further induce the
similarity of identical samples we apply the Triplet Hinge Loss detailed in
Section \ref{section:adaptive_margin}.

\subsection{Triplet Hinge Loss with Adaptive Margins}

\label{section:adaptive_margin}

The hinge loss is widely used in metric learning schemes as in Eqs.
\ref{eq:loss_pairwise} and \ref{equ:triplet loss}. It is common to set $m=1$
for $L_{2}$ normalized embeddings, and it provides adaptive Hard Negative
Mining, discarding negative samples with classification margins larger than
$m$, which are often uninformative.

We propose an adaptive margin $m$ that is updated throughout the training
iterations. Starting with a relatively small threshold $m,$ corresponding to a
weak separation of negative samples, and adaptively increasing $m$ as the
separation improves due to the CNN training.

We apply the adaptive margin to a symmetric Triplet Loss formulation as in Eq.
\ref{equ:triplet loss}
\begin{multline}
\mathcal{L}_{T}=\sum_{n=1}^{N}\left[  D(\mathbf{\hat{x}}_{n},\mathbf{\hat{y}%
}_{n}^{+})-D(\mathbf{\hat{x}}_{n},\mathbf{\hat{y}}_{n}^{-})+m_{x}\right]
_{+}\label{eq:triplet loss}\\
+\sum_{n=1}^{N}\left[  D(\widehat{\mathbf{y}}_{n},\mathbf{\hat{x}}_{n}%
^{+})-D(\widehat{\mathbf{y}}_{n},\mathbf{\hat{x}}_{n}^{-}))+m_{y}\right]  _{+}%
\end{multline}
where $\left\{  \mathbf{\hat{x}}_{n},\mathbf{\hat{y}}_{n}^{+},\mathbf{\hat{y}%
}_{n}^{-}\right\}  $ is a triplet of samples such that $\mathbf{\hat{x}}_{n}$
is an image embedding, while $\mathbf{\hat{y}}_{n}^{+}$ and $\mathbf{\hat{y}%
}_{n}^{-}$ are identical and non-identical captions, respectively, as given in
the training sets. The triplet $\left\{  \widehat{\mathbf{y}}_{n}%
,\mathbf{\hat{x}}_{n}^{+},\mathbf{\hat{x}}_{n}^{-}\right\}  $ is the symmetric
counterpart of the text embedding $\widehat{\mathbf{y}}_{n}$. $m_{x}$ and
$m_{y}$ are the adaptive margins for the image and text branches,
respectively, and $N$ is the number of negative and positive samples.

$m_{x}$ and $m_{y}$ are updated every $q$ batches, by computing the ratios
$M_{x}$ and $M_{y},$ of negative samples (within the batch) adhering to the
margins $m_{x}$ and $m_{y}$%
\begin{equation}
M_{x}=\frac{1}{N}\sum_{n=1}^{N}\#[D(\mathbf{\hat{x}}_{n},\mathbf{\hat{y}}%
_{n}^{+})-D(\mathbf{\hat{x}}_{n},\mathbf{\hat{y}}_{n}^{-})+m_{x}>0],
\label{eq:margin_ratio_x}%
\end{equation}%
\begin{equation}
M_{y}=\frac{1}{N}\sum_{n=1}^{N}\#[D(\widehat{\mathbf{y}}_{n},\mathbf{\hat{x}%
}_{n}^{+})-D(\widehat{\mathbf{y}}_{n},\mathbf{\hat{x}}_{n}^{-}))+m_{y}>0].
\label{eq:margin_ratio_y}%
\end{equation}

$M_{x}>r$ or $M_{y}>r$ implies that the CNN can separate a sufficient ratio of
the negative samples, and the margins $m_{x}$ and $m_{y}$ can be increased to
induce a stronger separation. Thus, the margins $m_{x}$ and $m_{y}$ are
updated by%
\begin{equation}
m_{x,y}=%
\begin{cases}
c\cdot m_{x,y} & \text{if $M_{x,y}>r$}\\
m_{x,y}, & \text{otherwise}.
\end{cases}
\label{eq:adaptive_margin_x}%
\end{equation}
\newline where $c>1$ is a predefined update multiplier.

Equations \ref{eq:margin_ratio_x}-\ref{eq:adaptive_margin_x} are applied every
$q$ batches to allows a statistically stable estimate of $M_{x,y}$.

\section{Experimental Results}

\label{sec:Experiments}

The proposed scheme was experimentally verified by applying it to multiple
contemporary image-text datasets used in previous state-of-the-art schemes. We
used the Flick8K \cite{HodoshFlickr8k}, Flickr30K \cite{flickr30k}, MS-COCO
\cite{mscoco} datasets, consisting of 8,000, 31,000 and 123,000 images,
respectively, where each image was annotated by five captions. Thus, for each
training dataset, we are given $\left\{  s_{n}\right\}  _{1}^{N}$ training
sets, each consisting of $s_{n}=\left\{  \mathbf{x}_{n},\mathbf{y}_{n}%
^{1},...,\mathbf{y}_{n}^{5}\right\}  $ $.$

In the training phase, the images were randomly resized in a scale of $\left[
0.8,1\right]  $, while the validation images were resized such that their
smallest dimension was $256$ pixels and their original aspect ratio was
retained. For the ResNet152 backbone, random crops of $224\times224$ pixels
were extracted and randomly flipped horizontally. In the testing phase,
following Huang et al. \cite{Huang2018LearningSC}, we applied 20 random crops
and averaged the resulting image embeddings. For the Attention-based VSRN
\cite{FRCNN} backbone\footnote{https://github.com/KunpengLi1994/VSRN}, we
applied the Faster RCNN to generate 36 ROIs per image in both training and
testing, and the images were augmented in the training phase by horizontal
flipping. Following \cite{SCAN,FRCNN}, the resulting VSRN-based model is an
ensemble computed by averaging the output embeddings of the three models which
performed best on the validation set. For both backbones, the mean of each
color channel was subtracted. In the text branch, the captions were tokenized
using the vocabulary provided by Faghri et al. \cite{vse++} such that each
sentence was encoded by a vector of tokens.

In the image annotation tasks, we computed the distance between the embedding
of the query image and the embeddings of all captions in the dataset. The
captions were ranked according to the minimal distance. Similarly, in the
image search task, we computed the distance between the embedding of the query
caption and the embeddings of all images in the dataset. The distance was then
used to rank the similarity.

As in previous works, we quantify the retrieval accuracy by R@K, which is
\textquotedblleft the recall at $K$\textquotedblright\ -- the fraction of
queries in which a correct item is retrieved within the $K$ embeddings closest
to the query. We consider two retrieval tasks. The first is \textit{image
annotation }(Img2Txt), where given a query image, we aim to retrieve the
corresponding captions, while the second is the \textit{image search}
(Txt2Img) task, where given a text query (caption) we search for the
corresponding image. Thus, an image annotation recall rate at $K$ (R@K), is
the average number of accurate captions retrieved, within the $K$ captions
nearest to the query image. Similarly, an image search recall rate of $K$
(R@K) is the average number of accurate images retrieved, given a text query,
within the $K$ most similar images.

\subsection{Training}

All CNN models were trained in two phases: first, computing the unquantized
embedding (Section \ref{section:constrained_loss}), and then refining the
resulting CNN to compute the quantized embedding (Section
\ref{section:quantized constrained_loss}). The \textit{unquantized} embedding
was computed using a precomputed image embedding CNN, where the succeeding
layers were trained by applying the Center Loss in Eq. \ref{equ:semantic loss}
and the adaptive hinge loss (Section \ref{section:adaptive_margin}). The
training samples $\left\{  s_{n}\right\}  _{1}^{N}$ were assigned to the
unquantized semantic centers $\left\{  \mathbf{c}_{n}\right\}  _{1}^{N}$, such
that $s_{n}$ is assigned to $\mathbf{c}_{n}$. The input to this training phase
is a batch of pairs of corresponding images and captions that are used as
positive pairs. The negative pairs are assigned by matching an image to
randomly chosen caption within the batch. The network is trained by optimizing
Eq. \ref{equ:loss}. The cross-entropy losses are trained using the labels of
the subsets $S=\left\{  \mathbf{s}_{n}\right\}  _{1}^{N}$, as in Section
\ref{chap:OurModel}. Thus, the image $\mathbf{x}_{n}$ and captions
$\mathbf{y}_{n}^{1},...,\mathbf{y}_{n}^{K}$ in a subset $\mathbf{s}%
_{n}=\left\{  \mathbf{x}_{n},\mathbf{y}_{n}^{1},...,\mathbf{y}_{n}%
^{K}\right\}  $ are labeled by $n$.

Next, we trained the \textit{quantized} embedding $C_{q}=\left\{
\widehat{\mathbf{c}}_{n}\right\}  _{1}^{N_{q}}$, by adding the
\textquotedblleft$N_{q\text{ }}$FC\textquotedblright\ and Softmax layers as in
Fig. \ref{fig:CNN outline}. As the layer \textquotedblleft$N_{q\text{ }}%
$FC\textquotedblright\ is uninitialized, it is first trained by freezing all
other CNN layers until convergence. We then unfroze all layers and fined-tuned
the network end-to-end. The quantized centers $\left\{  \widehat{\mathbf{c}%
}_{n}\right\}  _{1}^{N_{q}}$ were \textit{initialized} by applying KMEANS with
$N_{q}$ centers to $\left\{  \mathbf{c}_{n}\right\}  _{1}^{N}$. We also tried
using a random initialization of $\left\{  \widehat{\mathbf{c}}_{n}\right\}
_{1}^{N_{q}}$, and got similar results, but a longer convergence. The
cross-entropy losses in this phase, are as in the first
phase.\begin{table}[tbh]
\centering%
\begin{tabular}
[c]{|l|c|}\hline
\textbf{Layer} & \textbf{Output}\\\hline\hline
ResNet152 conv1 block & $112\times112\times64$\\
ResNet152 conv2.x block & $56\times56\times256$\\
ResNet152 conv3.x block & $28\times28\times512$\\
ResNet152 conv4.x block & $14\times14\times1024$\\
ResNet152 conv5.x block & $7\times7\times2048$\\
AveragePooling & $1\times2048$\\\hline
L2 Normalization & $1\times2048$\\
FC & $1\times1024$\\
L2 Normalization & $1\times1024$\\\hline
FC (softmax) & $1\times\#semantic$ $concepts$\\\hline
\end{tabular}
\caption{The layers of the image embedding branch corresponding to
$\mathbf{H_{i},}$ as depicted in Fig. \ref{fig:CNN outline}.}%
\label{table:image}%
\end{table}\begin{table}[tbh]
\centering%
\begin{tabular}
[c]{|l|c|}\hline
\textbf{Layer} & \textbf{Output}\\\hline\hline
Embedding & \#tokens $\times300$\\
Bi-Directional GRU & $1\times1024$\\
L2 Normalization & $1\times1024$\\\hline
FC (softmax) & $1\times\#semantic$ $concepts$\\\hline
\end{tabular}
\caption{The layers of the text embedding branch corresponding to
$\mathbf{H_{t},}$ as depicted in Fig. \ref{fig:CNN outline}.}%
\label{table:text}%
\end{table}

The Adam optimizer \cite{adam} was used to optimize the parameters of both
image and text branches with a varying learning rate starting at
$2\cdot10^{-4}$ for the first phase, and $2\cdot10^{-5}$ for the second and
third phases. The SGD optimizer was used to optimize the Semantic Center Loss
parameters using a learning rate of 0.5, where we reduced the learning rates
by a factor of $0.1$ every 15 epochs. We trained the first phase for 30
epochs, the second phase for an additional 15 epochs, and the third phase for
10 epochs. A fixed margin of $m_{x}=m_{y}=0.2$ was set for the hinge loss in
the first and second training phases, and both margins were optimized in the
third phase. We apply the proposed adaptive hinge loss to $\mathbf{\hat{x}}$
and $\mathbf{\hat{y}}$, as in Eq. \ref{eq:adaptive_margin_x}, by accumulating
the statistics $M_{x}$ and $M_{y}$ over $q=500$ iterations, and update the
margins $m_{x}$ and $m_{y}$ using the multiplier $c=1.03$, if more than
$r=0.8$ of the negative margins are beyond the hinge loss margins $M_{x}$ and
$M_{y}$. These settings were applied to training the models for all datasets.

As the proposed scheme can be used with any text and image embedding CNNs, we
applied it to two image-text matching CNNs. The first is detailed in Tables
\ref{table:image} and \ref{table:text} (ResNet-based), and the second is based
on the VSRN \cite{FRCNN} approach\footnote{Available at:
https://github.com/KunpengLi1994/VSRN}.

\subsection{Flickr30K Results}

The proposed scheme was also evaluated using the Flickr30K \cite{flickr30k}
dataset and was compared with multiple state-of-the-art schemes
\cite{vse++,dan,2WayNetwork,sm-lstm,emeddingNetwork,Huang2018LearningSC,SCAN,FRCNN}%
. The Flickr30K dataset consists of 31,000 Flickr images, each having five
captions, and following Karpathy and Fei-Fei \cite{feifeiKarpathy}, we used
1,000 images (and corresponding captions) for validation, 1,000 images for
testing and the rest for training. In the first phase of the training, we used
a batch size of 128 samples. We applied the proposed scheme using both the
VSRN \cite{FRCNN} and the CNN in Tables \ref{table:image} and \ref{table:text}%
, as a backbone for feature extraction. The ResNet-based results are reported
in the upper part of Table \ref{table:flick30k_results}, where the proposed
scheme implemented using 1000 quantization centers, outperforms all previous
schemes using the same backbone, in all categories, by an average of 3.3\%.
\begin{table}[tbh]
\centering%
\begin{tabular}
[c]{c|cccccc}\hline
Model & \multicolumn{3}{|c}{Annotation} & \multicolumn{3}{c}{Search}%
\\\cline{2-7}
& r@1 & r@5 & r@10 & \multicolumn{1}{|c}{r@1} & r@5 & r@10\\\hline
\multicolumn{7}{c}{VGG19}\\\hline\hline
\multicolumn{1}{l|}{Embedd\cite{emeddingNetwork}} & 40.7 & - & 79.2 & 29.2 &
- & 71.7\\
\multicolumn{1}{l|}{smLSTM\cite{sm-lstm}} & 42.5 & 71.9 & 81.5 & 30.2 & 60.4 &
72.3\\
\multicolumn{1}{l|}{2Way\cite{2WayNetwork}} & 49.8 & 67.5 & - & 36.0 & 55.6 &
-\\\hline
\multicolumn{7}{c}{ResNet152}\\\hline\hline
\multicolumn{1}{l|}{DAN\cite{dan}} & 55.0 & - & 89.0 & 39.4 & - & 79.1\\
\multicolumn{1}{l|}{VSE++\cite{vse++}} & 52.9 & 79.1 & 87.2 & 39.6 & 69.6 &
79.5\\
\multicolumn{1}{l|}{\textbf{Ours}} & \textbf{58.6} & \textbf{84.4} &
\textbf{91.1} & \textbf{45.5} & \textbf{75.2} & \textbf{83.9}\\\hline
\multicolumn{7}{c}{Attention-based}\\\hline\hline
\multicolumn{1}{l|}{Huang\cite{Huang2018LearningSC}} & 55.5 & 82.0 & 89.3 &
41.1 & 70.5 & 80.1\\
\multicolumn{1}{l|}{SCAN \cite{SCAN}} & 67.4 & 90.3 & 95.8 & 48.6 & 77.7 &
85.2\\
\multicolumn{1}{l|}{VSRN\cite{FRCNN}} & 71.3 & 90.6 & \textbf{96.0} & 54.7 &
81.8 & 88.2\\
\multicolumn{1}{l|}{\textbf{Ours}} & \textbf{73.3} & \textbf{92.2} &
\textbf{96.0} & \textbf{57.0} & \textbf{82.8} & \textbf{89.2}\\\hline
\end{tabular}
\caption{The Accuracy of the image annotation and search tasks evaluated using
the \textbf{Flickr30K dataset}. The center loss was applied to the image and
text embeddings as a semantic embedding. We also applied the Triplet Hinge
Loss with an Adaptive Margins, as in Section \ref{section:adaptive_margin}.}%
\label{table:flick30k_results}%
\end{table}

The lower part of Table \ref{table:flick30k_results} reports the results for
the SCAN \cite{SCAN} and VSRN \cite{FRCNN} schemes. Both of these schemes
utilize an ensemble consisting of two CNNs trained independently, where the
similarity results are averaged. Thus, in these schemes, multiple CNNs were
trained, and the optimal ensemble is chosen according to their joint
evaluation score on the validation set. We trained two CNNs based on the VSRN
backbone using 1000 quantization centers. It follows that each of our CNNs
outperforms the VSRN \cite{FRCNN} and SCAN \cite{SCAN} in 6/6 and
5/6\ categories, respectively. Training additional models from which to choose
the ensemble can improve our ensembles's accuracy.

\begin{figure}[tbh]
\centering{\ \includegraphics[width=1.0\linewidth]{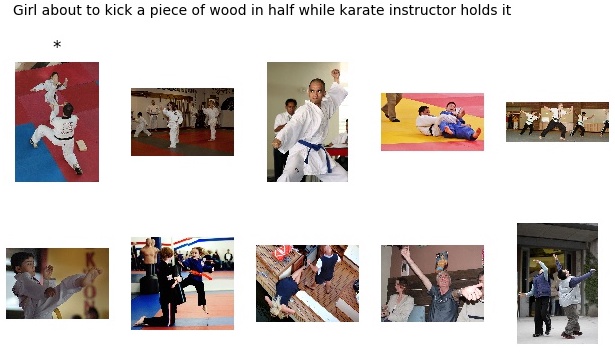}
}\caption{Image search (text \ to image retrieval) example taken from the
\textbf{Flickr30K} dataset. The query caption appears on top of the images.
The text embedding was compared to the embeddings of all of the images in the
dataset. The ten top ranked images are presented. The retrieved images are
depicted according to their similarity: left-to-right and then top-down. The
most similar image in on the upper-left corner, and is the correct retrieval
result.}%
\label{fig:f30k_ex_t2i}%
\end{figure}

\begin{figure}[tbh]
\centering{\ \includegraphics[width=0.95\linewidth]{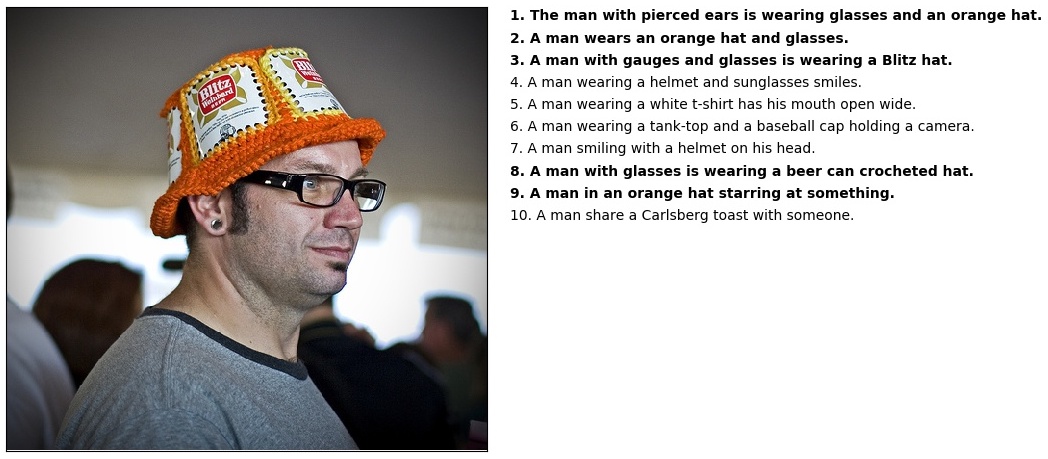}
}\caption{Image annotation (image to text retrieval) example taken from the
\textbf{Flickr30K} dataset. The image embedding was compared to the embeddings
of all captions in the dataset. The ten top ranked captions are presented, and
the correctly retrieved ones are marked in \textbf{bold}.}%
\label{fig:f30k_ex_i2t}%
\end{figure}

\begin{figure}[tbh]
\centering{\ \includegraphics[width=0.95\linewidth]{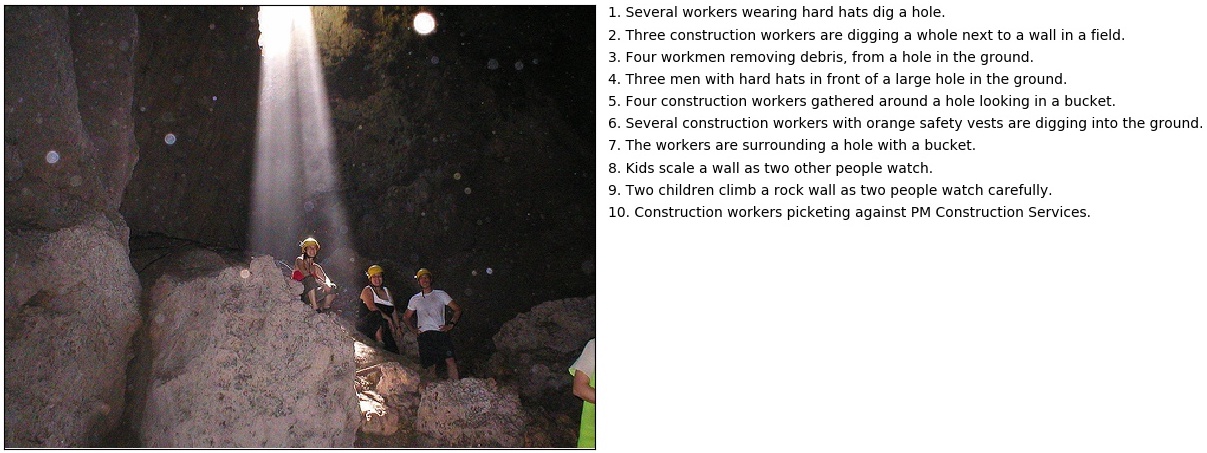}
}\caption{An example of a failed image annotation (image to text retrieval)
taken from the \textbf{Flickr30K} dataset. The top ten retrieval results bear
partial semantic similarity to the image content.}%
\label{fig:f30k_ex_i2t_failure}%
\end{figure}

Qualitative retrieval results are shown in Figs. \ref{fig:f30k_ex_t2i} and
\ref{fig:f30k_ex_i2t}, while an example of a failed image annotation task is
shown in Fig. \ref{fig:f30k_ex_i2t_failure}. These are the images whose
embedding is the closest to embedding of the query. By examining the failed
retrieval, it follows that although the retrieved captions are not part of the
groundtruth captions, are semantic similar to the query. Similarly, in Fig.
\ref{fig:f30k_ex_t2i}, the correct image is indeed retrieved first (shown on
the upper-left corner), but the following five retrieved images are also
related to the \textquotedblleft karate hat\textquotedblright\ query.


\subsection{MS-COCO Results}

\begin{table}[tbh]
\centering{%
\begin{tabular}
[c]{c|cccccc}\hline
Model & \multicolumn{3}{|c}{Annotation} & \multicolumn{3}{c}{Search}%
\\\cline{2-7}
& r@1 & r@5 & r@10 & \multicolumn{1}{|c}{r@1} & r@5 & r@10\\\hline
\multicolumn{7}{c}{VGG19}\\\hline\hline
\multicolumn{1}{l|}{Embedd\cite{emeddingNetwork}} & 50.4 & - & 69.4 & 39.8 &
- & 86.6\\
\multicolumn{1}{l|}{sm-LSTM\cite{sm-lstm}} & 53.2 & 83.1 & 91.5 & 40.7 &
75.8 & 87.4\\
\multicolumn{1}{l|}{2WayNet\cite{2WayNetwork}} & 55.8 & 75.2 & - & 39.7 &
63.3 & -\\\hline
\multicolumn{7}{c}{ResNet152}\\\hline\hline
\multicolumn{1}{l|}{VSE++\cite{vse++}} & 64.6 & 89.1 & 95.7 & 52.0 & 83.1 &
92.0\\
\multicolumn{1}{l|}{\textbf{Ours}} & \textbf{69.3} & \textbf{93.2} &
\textbf{97.7} & \textbf{56.0} & \textbf{86.6} & \textbf{93.7}\\\hline
\multicolumn{7}{c}{Attention-based}\\\hline\hline
\multicolumn{1}{l|}{Huang\cite{Huang2018LearningSC}} & 69.9 & 92.9 & 97.5 &
56.7 & 87.5 & 94.8\\
\multicolumn{1}{l|}{SCAN \cite{SCAN}} & 72.7 & 94.8 & 98.4 & 58.8 & 88.4 &
94.8\\
\multicolumn{1}{l|}{VSRN \cite{FRCNN}} & 76.2 & 94.8 & 98.2 & 62.8 & 89.7 &
95.1\\
\multicolumn{1}{l|}{\textbf{Ours}} & \textbf{76.6} & \textbf{95.6} &
\textbf{98.6} & \textbf{63.9} & \textbf{90.3} & \textbf{95.4}\\\hline
\end{tabular}
}\caption{The Accuracy of the image annotation and search tasks evaluated
using the \textbf{MS-COCO dataset 1K test}. The center loss was applied to the
image and text embeddings as a semantic embedding.}%
\label{table:coco_results1k}%
\end{table}The MS-COCO \cite{mscoco} dataset is significantly larger than the
Flickr8K and Flickr30K datasets and was split to 5000 images for test, 5000
images for validation, while the remaining 113,287 images were used for
training, following Karpathy and Fei-Fei \cite{feifeiKarpathy}. In the first
phase of the training, we used a batch size of 128, while in the second and
third phases, we applied a batch size of 64. Due to the large number of
training images, we applied the quantized semantic loss introduced in Section
\ref{section:quantized constrained_loss}, using 1000 quantized semantic
centers. We report the results of two test setups:\ In the first, reported in
Table \ref{table:coco_results1k}, the 5000 test images were split into five
sets of 1000 images (and captions), and the results are averaged over the five
folds (COCO1K). In the second, reported in Table \ref{table:coco_results5k},
the results of the 5000 test images are averaged as a single test set
(COCO5K). Same as in the previous datasets, we trained different CNNs using
the VSRN \cite{FRCNN} and ResNet backbones. For the COCO1K, we outperform the
SCAN and VSRN schemes in \textit{all} categories. For the COCO5K dataset, our
approach outperformed the previous best ResNet-based approaches
(Huang\cite{Huang2018LearningSC}) in 5/6 categories, by an average of 1.5\%.
Comparing the accuracy for the Flickr8K, Flickr30K and MS-COCO in Tables
\ref{table:flick8k}, \ref{table:flick30k_results} and
\ref{table:coco_results5k}, respectively, it follows that training the
proposed scheme on larger datasets consistently improves the retrieval
accuracy.\begin{table}[tbh]
\centering{%
\begin{tabular}
[c]{c|cccccc}\hline
Model & \multicolumn{3}{|c}{Annotation} & \multicolumn{3}{c}{Search}%
\\\cline{2-7}
& r@1 & r@5 & r@10 & \multicolumn{1}{|c}{r@1} & r@5 & r@10\\\hline
\multicolumn{7}{c}{VGG19}\\\hline\hline
\multicolumn{1}{l|}{Embedd\cite{emeddingNetwork}} & 50.4 & - & 69.4 & 39.8 &
- & 86.6\\
\multicolumn{1}{l|}{smLSTM\cite{sm-lstm}} & 53.2 & 83.1 & 91.5 & 40.7 & 75.8 &
87.4\\
\multicolumn{1}{l|}{2WayNet\cite{2WayNetwork}} & 55.8 & 75.2 & - & 39.7 &
63.3 & -\\\hline
\multicolumn{7}{c}{ResNet152}\\\hline\hline
\multicolumn{1}{l|}{VSE++\cite{vse++}} & 41.3 & 69.2 & 81.2 & 30.3 & 59.1 &
72.4\\
\multicolumn{1}{l|}{Huang\cite{Huang2018LearningSC}} & 42.8 & 72.3 & 83.0 &
\textbf{33.1} & 62.9 & 75.5\\
\multicolumn{1}{l|}{\textbf{Ours}} & \textbf{44.7} & \textbf{75.4} &
\textbf{86.0} & \textbf{33.1} & \textbf{63.7} & \textbf{75.8}\\\hline
\end{tabular}
}\caption{The Accuracy of the image annotation and search tasks evaluated
using the \textbf{MS-COCO dataset 5K test}. The center loss was applied to the
image and text embeddings as a semantic embedding.}%
\label{table:coco_results5k}%
\end{table}

\begin{figure}[tbh]
\centering{\ \includegraphics[width=0.90\linewidth]{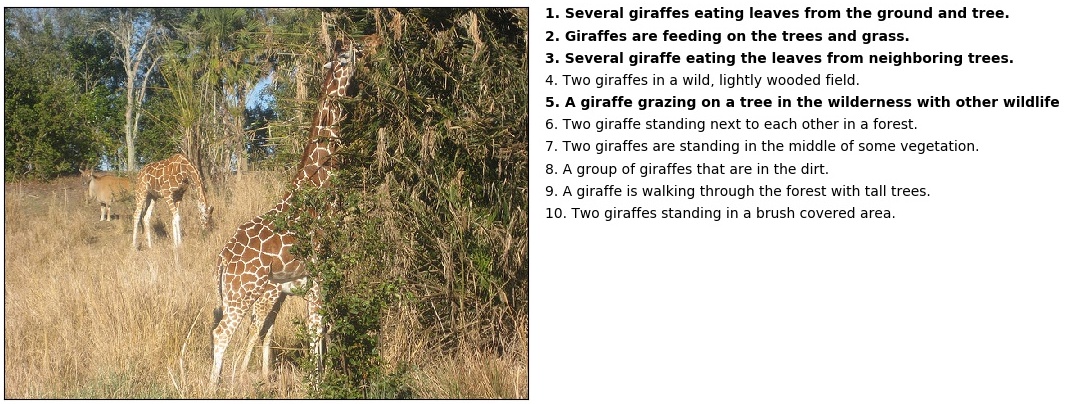}
}\caption{Image annotation (image to text retrieval) example taken from the
\textbf{MS-COCO} dataset. The image embedding was compared to the embeddings
of all captions in the dataset. The ten top ranked captions are presented, and
the correctly retrieved ones are marked in \textbf{bold}.}%
\label{fig:coco_ex_i2t}%
\end{figure}

\begin{figure}[tbh]
\centering{\ \includegraphics[width=0.90\linewidth]{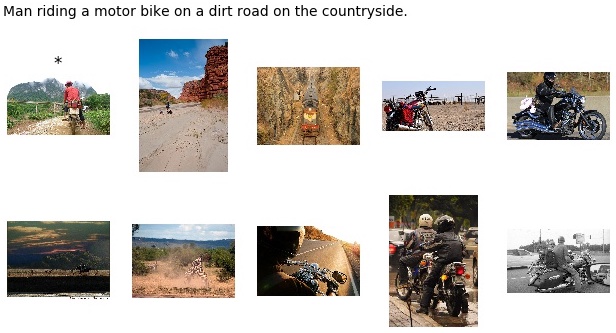}}%
\caption{Image search (text \ to image retrieval) example taken from the
\textbf{MS-COCO} dataset. The query caption appears on top of the images. The
text embedding was compared to the embeddings of all of the images in the
dataset. The ten top ranked images are presented. The retrieved images are
depicted according to their similarity: left-to-right and then top-down. The
most similar image in on the upper-left corner, and is the correct result. }%
\label{fig:coco_ex_t2i}%
\end{figure}

\begin{figure}[tbh]
\centering{\ \includegraphics[width=0.90\linewidth]{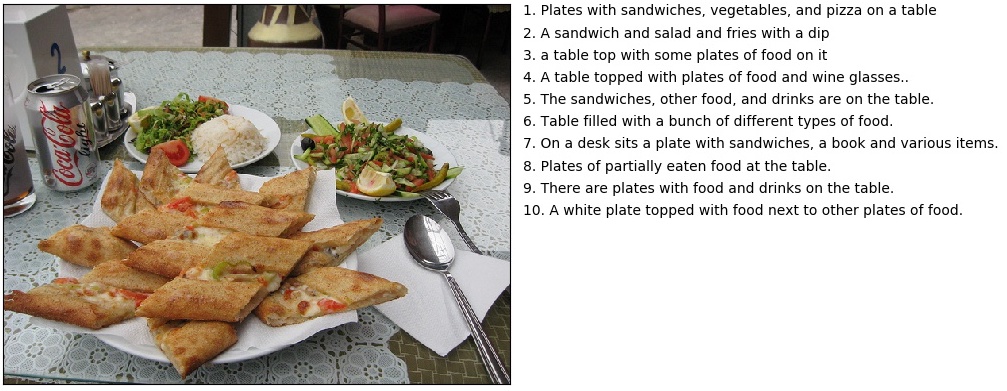}
}\caption{An example of failed image annotation (image to text retrieval)
taken from the \textbf{MS-COCO} dataset. The top ten retrieval results bear
partial semantic similarity to the image content.}%
\label{fig:coco_ex_i2t_failure}%
\end{figure}

Qualitative retrieval results are shown in Figs. \ref{fig:coco_ex_i2t} and
\ref{fig:coco_ex_t2i}, while an image annotation failure is shown in Fig.
\ref{fig:coco_ex_i2t_failure}.

\subsection{Flickr8K Results}

\begin{figure}[tbh]
{\ \includegraphics[width=1.0\linewidth]{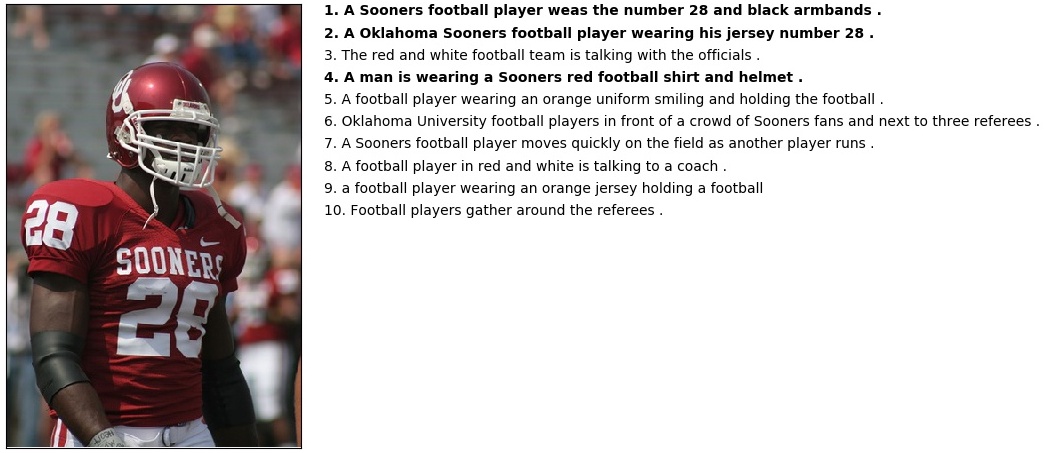}
}\caption{Image annotation (image to text retrieval) example taken from the
\textbf{Flickr8K} dataset. The image embedding was compared to the embeddings
of all captions in the dataset. The ten top ranked captions are presented, and
the correctly retrieved ones are marked in \textbf{bold}.}%
\label{fig:f8k_ex_i2t}%
\end{figure}

\begin{figure}[tbh]
{\includegraphics[width=1.0\linewidth]{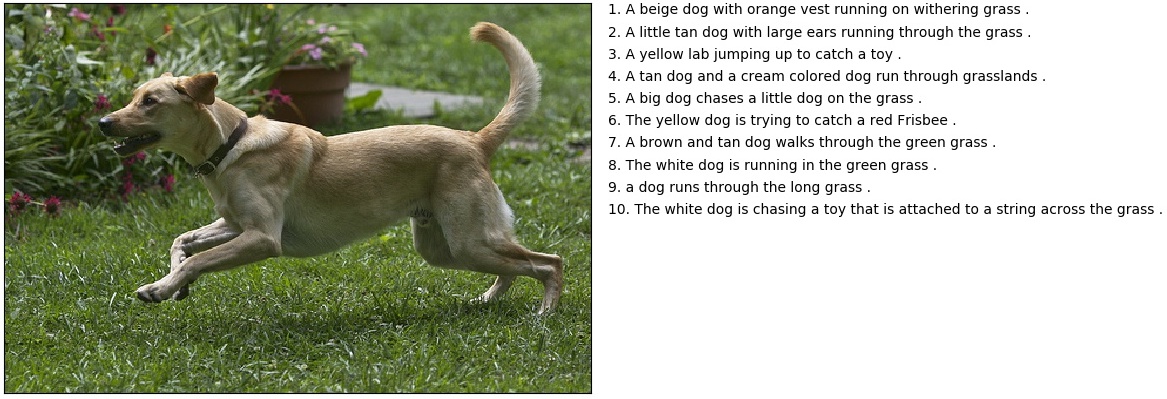}
}\caption{An example of failed image annotation (image to text retrieval)
taken from the \textbf{Flickr8K} dataset. The top ten retrieval results bear
partial semantic similarity to the image content.}%
\label{fig:f8k_ex_i2t_failure}%
\end{figure}

\begin{figure}[tbh]
{ \includegraphics[width=1.0\linewidth]{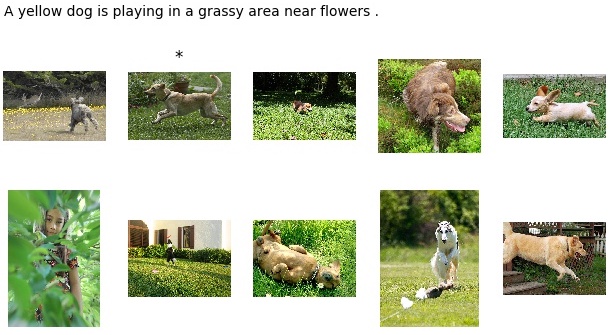}
}\caption{Image search (text\ to image retrieval) example taken from the
\textbf{Flickr8K} dataset. The query caption appears on top of the images. The
text embedding was compared to the embeddings of all of the images in the
dataset. The ten top ranked images are presented. The retrieved images are
depicted according to their similarity: left-to-right and then top-down. The
reference result is marked by *. It seems that the leading retrieval result
(top-left) corresponds to the given caption as well.}%
\label{fig:f8k_ex_t2i}%
\end{figure}

\begin{table}[tbh]
\centering%
\begin{tabular}
[c]{c|cccc}\hline
Model & \multicolumn{2}{|c}{Annotation} & \multicolumn{2}{|c}{Search}%
\\\cline{2-5}
& r@1 & r@5 & \multicolumn{1}{|c}{r@1} & r@5\\\hline\hline
\multicolumn{1}{l|}{CCA \cite{cca}} & 31.0 & 59.3 & 21.3 & 50.1\\
\multicolumn{1}{l|}{RNN-FV \cite{rnn-fv}} & 31.6 & 61.2 & 23.2 & 53.3\\
\multicolumn{1}{l|}{VQA-A \cite{vqa-a}} & 24.3 & 52.2 & 17.2 & 42.8\\
\multicolumn{1}{l|}{2WayNet \cite{2WayNetwork}} & 43.4 & 63.2 & 29.3 & 49.7\\
\multicolumn{1}{l|}{\textbf{Ours} 50c} & 38.9 & 67.9 & 29.2 & 59.6\\
\multicolumn{1}{l|}{\textbf{Ours} 100c} & \textbf{44.1} & \textbf{69.7} &
\textbf{29.9} & 59.7\\
\multicolumn{1}{l|}{\textbf{Ours} 1000c} & 39.2 & 67.8 & 29.4 & 59.3\\
\multicolumn{1}{l|}{\textbf{Ours} 6000c} & 38.6 & 68.3 & \textbf{29.9} &
\textbf{60.3}\\\hline
\multicolumn{5}{c}{VSRN\cite{FRCNN}}\\\hline\hline
\multicolumn{1}{l|}{VSRN} & 42.5 & 73.9 & 33.5 & 64.2\\
\multicolumn{1}{l|}{\textbf{Ours} 1000c} & \textbf{42.9} & \textbf{74.3} &
\textbf{34.3} & \textbf{64.7}\\\hline
\end{tabular}
\caption{The accuracy of the image annotation and search tasks evaluated using
the \textbf{Flickr8K dataset}. The center loss was applied to the image and
text embeddings as a semantic embedding. We also applied the Triplet Hinge
Loss with Adaptive Margins, as in Section \ref{section:adaptive_margin}.}%
\label{table:flick8k}%
\end{table}We applied the proposed scheme to the Flickr8K
\cite{HodoshFlickr8k} dataset, which was split to 1000 images (and
corresponding captions) for test, 1000 images for validation, and 6000 images
for training, following Karpathy and Fei-Fei \cite{feifeiKarpathy}. We used a
batch size of 128, and applied the proposed semantic embedding to two
backbones: the one in Tables \ref{table:image}, \ref{table:text}, and the one
based on the VSRN \cite{FRCNN} approach. The retrieval results are reported in
Table \ref{table:flick8k}, where we separately compare the results of CNN
trained with different backbones. It follows that the proposed approach
outperforms all previous schemes in general, and improves over the
VSRN\cite{FRCNN} and other backbones by an average of 0.6\% and 4.5\% ,
respectively. As for the number of centers, we tested several configurations
and the `sweet spot' of 100 centers performs best, on both validation and test
sets. Qualitative examples are shown in Figs. \ref{fig:f8k_ex_i2t} and
\ref{fig:f8k_ex_t2i}, where an image annotation failure is shown in Fig.
\ref{fig:f8k_ex_i2t_failure}. We note that although the caption retrieved
first in Fig. \ref{fig:f8k_ex_i2t_failure} is incorrect, it seems to adhere to
the caption.

\subsection{Ablation study}

\label{section:combinations}

We conducted an ablation study to evaluate the influence and importance of the
multiple algorithmic components of the proposed scheme. We first studied the
use of the Cross Entropy losses (as in Fig. \ref{fig:CNN outline}) and the
proposed Semantic Centers formulation. For that, we applied different
combinations of these losses to the Flickr8K and Flickr30K datasets. The
Center loss was applied in an unquantized formulation, and the results are
shown in Tables \ref{table:center_crossentropy_analysis}. Using the Semantic
Center loss (without the Cross Entropy losses) provides an average accuracy
improvement of 19.3\% and 2.5\%, when applied to the Flickr8K and Flickr30K
datasets, respectively, compared to applying the Center loss to the text
embeddings only. In contrast, applying the Cross Entropy losses, alongside the
semantic losses, provides an average improvement of only 0.3\%. Hence, the
proposed Semantic Center loss is the one doing the heavy lifting in our
scheme.\begin{table}[tbh]
\centering%
\begin{tabular}
[c]{l|c|cccc}\hline
\multicolumn{2}{c|}{\textbf{Model}} & \multicolumn{2}{|c}{\textbf{Annotation}}
& \multicolumn{2}{|c}{\textbf{Search}}\\\hline
\textbf{Centers} & \textbf{CE loss} & r@1 & r@10 & \multicolumn{1}{|c}{r@1} &
r@10\\\hline\hline
\multicolumn{6}{c}{Flickr8K}\\\hline\hline
Txt & - & 6.6 & 33.9 & 15.7 & 53.6\\
Txt & Txt & 9.7 & 42.1 & 15.3 & 53.8\\
Semantic & - & 30.1 & \textbf{73.6} & 21.6 & 61.9\\
Semantic & Img/Txt & \textbf{32.3} & 71.6 & \textbf{22.1} & \textbf{63.4}%
\\\hline
\multicolumn{6}{c}{Flickr30K}\\\hline\hline
Txt & - & 43.9 & 81.2 & 30.9 & 71.9\\
Txt & Txt & 45.1 & 82.6 & 32.5 & 72.9\\
Semantic & - & 46.7 & 83.4 & \textbf{33.7} & \textbf{74.0}\\
\multicolumn{1}{l|}{Semantic} & Img/Txt & \textbf{46.9} & \textbf{84.0} &
33.3 & 73.8\\\hline
\end{tabular}
\caption{Ablation study of the Semantic Centers, and the Cross-Entropy (CE)
losses, using the Flickr8K and Flickr30K datasets. The Center loss is applied
to either the text embeddings (txt), or as Semantic centers (Semantic) to both
the text and image embeddings.}%
\label{table:center_crossentropy_analysis}%
\end{table}

In Table \ref{table:center_crossentropy_analysis_f30k_resnet_backbone}, we
compared using the unquantized and quantized Semantic Center losses (with 1000
centers) to a baseline without the Center Loss. For that we applied the
Resnet152 \cite{dan,vse++} and VSRN \cite{FRCNN} backbones to the Flickr30K
dataset. For the Resnet152 backbone, without using the random crops, the
Center loss outperformed the baseline by $2.5\%$. In contrast, when random
crops were used, the average improvement of the Center loss was only $0.8\%$.
Using the random crops improved the accuracy by 2\%. In contrast, applying the
VSRN backbone improves the overall average accuracy by a significant $6\%$ and
the average improvement of the Center loss is $0.8\%$. It seems that as the
backbone CNN becomes more elaborate (VSRN vs. Resnet152), the available margin
for improvement reduces. Still, the improvement achieved by the Center loss is
complementary to improving the backbone.\begin{table}[tbh]
\centering%
\begin{tabular}
[c]{c|cccccc}\hline
Center Loss & \multicolumn{3}{|c}{Annotation} & \multicolumn{3}{c}{Search}%
\\\cline{2-7}
& r@1 & r@5 & r@10 & \multicolumn{1}{|c}{r@1} & r@5 & r@10\\\hline
\multicolumn{7}{c}{Resnet152 backbone - No Random Crops}\\\hline\hline
\multicolumn{1}{l|}{No} & 52.9 & 79.1 & 87.2 & 39.6 & 69.6 & 79.5\\
\multicolumn{1}{l|}{Unquantized} & 55.8 & 82.5 & \textbf{89.3} & 41.8 & 71.5 &
79.9\\
\multicolumn{1}{l|}{Quantized} & \textbf{57.2} & \textbf{83.6} & \textbf{89.3}
& \textbf{42.3} & \textbf{72.0} & \textbf{80.3}\\\hline
\multicolumn{7}{c}{Resnet152 backbone - With Random Crops}\\\hline\hline
\multicolumn{1}{l|}{No} & 57.1 & 85.2 & 91.5 & 41.9 & 72.3 & 82.2\\
\multicolumn{1}{l|}{Unquantized} & 59.9 & 83.9 & 90.8 & 43.9 & 73.5 & 81.8\\
\multicolumn{1}{l|}{Quantized} & \textbf{60.1} & \textbf{83.7} & \textbf{91.9}
& \textbf{44.6} & \textbf{73.9} & \textbf{82.5}\\\hline
\multicolumn{7}{c}{VSRN backbone- No ensemble}\\\hline\hline
\multicolumn{1}{l|}{No} & 68.1 & 89.5 & 94.2 & 50.5 & 78.8 & 86.4\\
\multicolumn{1}{l|}{Unquantized} & 68.3 & \textbf{90.4} & \textbf{94.9} &
51.9 & \textbf{80.2} & 87.0\\
\multicolumn{1}{l|}{Quantized} & \textbf{69.8} & 88.7 & 94.6 & \textbf{52.2} &
79.8 & \textbf{87.0}\\\hline
\end{tabular}
\caption{An ablation study of the configurations of the Semantic Center loss.
The schemes were applied to the Flickr30K dataset, using 1000 quantized
Semantic centers.}%
\label{table:center_crossentropy_analysis_f30k_resnet_backbone}%
\end{table}

The sensitivity to the number of Semantic Centers is studied in Table
\ref{table:center_f30k_centers_no}, where we applied varying numbers of
centers to the Flickr8K, Flickr30K and MS-COCO1K datasets. We computed the
maximal difference in average accuracy, that was $1\%$ for the Flickr8K, in
contrast to $0.1\%$ for the Flickr30K and MS-COCO1K datasets. This implies
that the choice of the number of centers is robust.\begin{table}[tbh]
\centering%
\begin{tabular}
[c]{c|cccccc}\hline
\#Centers & \multicolumn{3}{|c}{Annotation} & \multicolumn{3}{c}{Search}%
\\\cline{2-7}
& r@1 & r@5 & r@10 & \multicolumn{1}{|c}{r@1} & r@5 & r@10\\\hline
\multicolumn{7}{c}{Flickr 8K}\\\hline\hline
\multicolumn{1}{l|}{50} & 38.9 & 67.9 & - & 29.2 & 59.6 & -\\
\multicolumn{1}{l|}{100} & \textbf{44.1} & \textbf{69.7} & - & \textbf{29.9} &
\textbf{59.7} & -\\
\multicolumn{1}{l|}{1000} & 39.2 & 67.8 & - & 29.4 & 59.3 & -\\\hline
\multicolumn{7}{c}{Flickr 30K}\\\hline\hline
\multicolumn{1}{l|}{100} & 56.7 & 83.4 & 89.8 & 42.1 & 71.9 & 80.3\\
\multicolumn{1}{l|}{500} & \textbf{57.8} & 82.9 & 89.8 & 42.0 & \textbf{72.1}
& 80.2\\
\multicolumn{1}{l|}{1000} & 57.2 & \textbf{83.6} & 89.3 & \textbf{42.3} &
72.0 & \textbf{80.3}\\
\multicolumn{1}{l|}{5000} & 57.3 & 82.9 & \textbf{90.0} & 41.8 & 71.9 &
80.0\\\hline
\multicolumn{7}{c}{MS-COCO 1K}\\\hline\hline
\multicolumn{1}{l|}{500} & 66.5 & \textbf{91.6} & \textbf{96.7} & 53.8 &
85.0 & \textbf{92.6}\\
\multicolumn{1}{l|}{1000} & 66.5 & 91.4 & 96.6 & \textbf{53.9} & \textbf{85.1}
& \textbf{92.6}\\
\multicolumn{1}{l|}{5000} & \textbf{66.7} & 91.2 & \textbf{96.7} &
\textbf{53.9} & 85.0 & \textbf{92.6}\\\hline
\end{tabular}
\caption{Ablation study results for number of centers for the quantization
phase. We used a ResNet152 backbone for these results, and the Flickr 30k and
MS-COCO were tested without random crops.}%
\label{table:center_f30k_centers_no}%
\end{table}

We also studied using a Transformer-based text embedding, instead of the
embedding layer-based embedding used in previous works
\cite{vse++,Huang2018LearningSC,SCAN,FRCNN}. For that we applied the
pretrained \textquotedblleft bert-large-uncased\textquotedblright\ model by
Huggingface\footnote{https://huggingface.co/bert-large-uncased}. The model was
fine-tuned by setting its learning-rate to $10^{-5}$. The results are shown in
Table \ref{table:text transformer}, where the use of the Bert-based language
model improved the retrieval accuracy by an average of 1\%. This is an
insignificant improvement considering the huge number of parameters (336M
parameters). It seems that due to the relatively limited vocabulary of the
captions and their short length, a simple language model based on learnt
embeddings is able to perform well.\begin{table}[tbh]
\centering%
\begin{tabular}
[c]{c|cccccc}\hline
Text embedding & \multicolumn{3}{|c}{Annotation} & \multicolumn{3}{c}{Search}%
\\\cline{2-7}
& r@1 & r@5 & r@10 & \multicolumn{1}{|c}{r@1} & r@5 & r@10\\\hline
\multicolumn{7}{c}{Resnet152 backbone - With Random Crops}\\
\multicolumn{1}{l|}{Embedding} & 60.1 & 83.7 & 91.9 & 44.6 & 73.9 & 82.5\\
\multicolumn{1}{l|}{Transformer} & \textbf{60.4} & \textbf{84.2} &
\textbf{92.8} & \textbf{45.5} & \textbf{74.1} & \textbf{83.4}\\\hline
\multicolumn{7}{c}{VSRN backbone- No ensemble}\\\hline\hline
\multicolumn{1}{l|}{Embedding} & 69.8 & 88.7 & 94.6 & 52.2 & 79.8 & 87.0\\
\multicolumn{1}{l|}{Transformer} & \textbf{70.6} & \textbf{89.1} &
\textbf{95.4} & \textbf{52.3} & \textbf{80.2} & \textbf{87.9}\\\hline
\end{tabular}
\caption{An ablation study of the text embedding. We compare using learnt
embeddings and the \textquotedblleft bert-large-uncased\textquotedblright%
\ Transformer-based languge\ model by Huggingface. The schemes were applied to
the Flickr30K dataset, using 1000 quantized Semantic centers.}%
\label{table:text transformer}%
\end{table}

We also examined the effectiveness of the proposed adaptive margin scheme
introduced in Section \ref{section:adaptive_margin}. For that we applied the
proposed scheme to the Flickr30K dataset using different values of the scaling
factor $c$ defined in Eq. \ref{eq:adaptive_margin_x}. We also compare to using
the adaptive margins proposed by Wu et al. \cite{Sampling-Matters} and Li
\cite{Li_2020_CVPR}. The choice of $c$ seems to be robust. Choosing
$c\in\left[  1.03,1.05\right]  $ provides an accuracy improvement of 0.7\%,
while outperforming the Adaptive Margin schemes by Wu \cite{Sampling-Matters}
and Li \cite{Li_2020_CVPR} by close to $25\%$ and $15\%$,
respectively.\begin{table}[tbh]
\centering%
\begin{tabular}
[c]{c|c|ccccccc}\hline
Adaptive &  & \multicolumn{3}{|c}{Annotation} & \multicolumn{3}{c}{Search} &
\\\cline{2-9}%
Margin & $c$ & r@1 & r@5 & r@10 & \multicolumn{1}{|c}{r@1} & r@5 & r@10 &
Avg.\\\hline
\multicolumn{1}{l|}{Non} & - & 55.8 & 82.5 & 89.3 & 41.8 & 71.5 & 79.9 &
70.1\\
\multicolumn{1}{l|}{\textbf{Our }} & 1.10 & 55.9 & 80.6 & 89.2 & 41.9 & 71.8 &
\textbf{80.6} & 70.0\\
\multicolumn{1}{l|}{\textbf{Our}} & 1.05 & \textbf{57.7} & \textbf{82.8} &
89.1 & 41.4 & 70.9 & 80.2 & 70.4\\
\multicolumn{1}{l|}{\textbf{Our}} & 1.03 & 57.0 & 82.7 & \textbf{89.9} &
\textbf{42.1} & \textbf{71.8} & 80.5 & \textbf{70.7}\\
\multicolumn{1}{l|}{Wu \cite{Sampling-Matters}} & - & 28.6 & 54.8 & 65.8 &
19.1 & 44.1 & 54.9 & 44.6\\
\multicolumn{1}{l|}{Li \cite{Li_2020_CVPR}} & - & 47.6 & 69.3 & 75.8 & 24.6 &
53.7 & 64.4 & 55.9\\\hline
\end{tabular}
\caption{Ablation study of the proposed adaptive margin, evaluated using the
Flickr30K dataset. We compare with the adaptive margins scheme by Wu et al.
\cite{Sampling-Matters}.}%
\label{table:center_crossentropy_analysis_f30k_adaptive_margin}%
\end{table}

\section{Conclusions}

\label{sec:Conclusions}

In this work, we proposed an approach for computing joint image-text
embeddings that encode the semantic similarity in a Euclidean $L_{2}$ space.
For that, we first applied the semantic center loss to encode
\textit{identical} semantic concepts based on the training set. By introducing
a differentiable quantization into the end-to-end trainable network, we derive
an embedding that encodes semantic \textit{similarity}. The similarity is thus
encoded automatically, without having to define the similar concepts
explicitly, based on textual analysis, as in previous works. We also propose a
novel formulation of the triplet loss using adaptive hinge loss margins that
are updated continuously throughout the training, allowing improved retrieval
accuracy. The proposed contributions can be applied to \textit{any} CNN
subnetworks for text and image embedding such as the ResNet152 CNN and GRU, or
VSRN \cite{FRCNN} used in this implementation. Our scheme is experimentally
shown to compare favorably with contemporary state-of-the-art approaches when
applied to multiple image-text datasets.

\bibliographystyle{IEEEtran}
\bibliography{citations}

\begin{IEEEbiography}[{\includegraphics[width=1in,height=1.25in,clip,keepaspectratio]{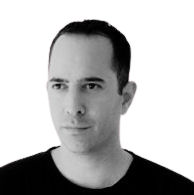}}]{Noam Malali}
earned a B.Sc. and an M.Sc. in electrical engineering from Bar-Ilan University in 2011 and 2020, respectively. His research interests center on deep learning and machine learning. Expert in applied machine learning, he has broad experience in many areas of research in the hi-tech industry and academia.
\end{IEEEbiography}

\begin{IEEEbiography}[{\includegraphics[width=1in,height=1.25in,clip,keepaspectratio]{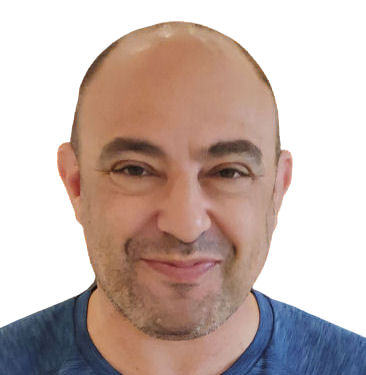}}]{Yosi Keller}
Yosi earned his BSc in electrical engineering from Technion-Israel Institute of Technology, Haifa, in 1994, and his MSc and Ph.D. degrees in electrical engineering from Tel Aviv University, summa cum laude, in 1998 and 2003, respectively. He was a Gibbs Assistant Professor at Yale University from 2003 to 2006. He is an Associate Professor at the Faculty of Engineering at Bar Ilan University in Ramat-Gan, Israel. His research interests are in Computer Vision, Deep Learning, and Biometrics.
\end{IEEEbiography}

\end{document}